\pdfoutput=1
\documentclass[letterpaper, 10pt, conference]{ieeeconf}      % Use this line for a4
\usepackage{amssymb,gensymb,graphicx,amsmath,color,algorithm,algorithmic,tabularx,caption,subcaption,float,adjustbox,todonotes}
\usepackage{balance}
\usepackage{dblfloatfix}
\usepackage[noadjust]{cite} % groups citations
\usepackage{cleveref}

\graphicspath{ {./figures/} }

\IEEEoverridecommandlockouts                              % This command is only
\title{\LARGE \bf
NeuroTac: A Neuromorphic Optical Tactile Sensor \\ applied to Texture Recognition
\vspace{0em}
}

\author{Benjamin Ward-Cherrier, {\em Member, IEEE}, Nicholas Pestell, {\em Student Member, IEEE},\\ 
	 Nathan F. Lepora, {\em Member, IEEE} % <-this % stops a space
\thanks{BWC was supported by a University of Bristol Vice-Chancellor's fellowship, NP was supported by an EPSRC DTP studentship and NL was supported in part by a Leverhulme Trust Research Leadership Award on 'A biomimetic forebrain for robot touch' (RL-2016-039).}% <-this % stops a space
\thanks{Authors are with the Department of Engineering Mathematics, University of Bristol and Bristol Robotics Laboratory, University of Bristol, UK.\newline Email: \{b.ward-cherrier, n.pestell , n.lepora\}@bristol.ac.uk}}
\begin{document}

\maketitle 

\thispagestyle{empty}
\pagestyle{empty}

%%%%%%%%%%%%%%%%%%%%%%%%%%%%%%%%%%%%%%%%%%%%%%%%%%%%%%%%%%%%%%%%%%%%%%%%%%%%%%%%

\begin{abstract}
Developing artificial tactile sensing capabilities that rival human touch is a long-term goal in robotics and prosthetics. Gradually more elaborate biomimetic tactile sensors are being developed and applied to grasping and manipulation tasks to help achieve this goal. Here we present the neuroTac, a novel neuromorphic optical tactile sensor. The neuroTac combines the biomimetic hardware design from the TacTip sensor which mimicks the layered papillae structure of human glabrous skin, with an event-based camera (DAVIS240, iniVation) and algorithms which transduce contact information in the form of spike trains. The performance of the sensor is evaluated on a texture classification task, with four spike coding methods being implemented and compared: Intensive, Spatial, Temporal and Spatiotemporal. We found timing-based coding methods performed with the highest accuracy over both artificial and natural textures. The spike-based output of the neuroTac could enable the development of biomimetic tactile perception algorithms in robotics as well as non-invasive and invasive haptic feedback methods in prosthetics.

\end{abstract}

%%%%%%%%%%%%%%%%%%%%%%%%%%%%%%%%%%%%%%%%%%%%%%%%%%%%%%%%%%%%%%%%%%%%%%%%%%%%%%%%%%%%%%%%%%%%%%%%%%%%%%%%
\section{INTRODUCTION}

%The sense of touch in humans allows us to interact with our environment, and is an essential component of both human social interaction~\cite{dunbar2010social}, and in-hand object manipulation~\cite{yousef2011tactile}. 

The long-term scientific goal of human-like artificial touch is being worked towards with the creation of gradually more elaborate biomimetic tactile sensors. The development of sensors which mimic aspects of biological touch could lead to safer robots and more intuitive and versatile prosthetic devices. An example of such sensors are neuromorphic tactile sensors, which aim to replicate the spike-based representation of information found in the nervous system. 

The principal objective of neuromorphic engineering is to develop technologies which can exploit efficient representations of information and operate in a rapid, power-efficient manner. Using neuromorphic technologies with event-based outputs also holds potential for integration with the human nervous system, as has been demonstrated with artificial retinas~\cite{zaghloul2006silicon}. Optical tactile sensors have also recently demonstrated progress on a number of tactile tasks~\cite{li2013sensing,james2018slip}, and have the advantage of capitalizing on advances in image recognition and machine learning techniques. These technologies will be combined here to develop a tactile sensor for robotic manipulation and prosthetics applications.

The aims of this paper are as follows: 
\begin{itemize}
\item Develop a novel neuromorphic optical tactile sensor to enable the development and investigation of biomimetic spike-based information processing methods. 
\item Validate this sensor on a texture classification task.
\item Investigate 4 spike coding methods (intensive, spatial, temporal and spatiotemporal) and their effect on texture recognition performance.
\end{itemize}

%\notes{Artificial tactile sensors have been developed with a variety of underlying transduction methods and technologies~\cite{dahiya2010tactile}. Each method will generally have its own advantages and disadvantages, making it best suited for particular tasks or applications. Here, we explore a novel type of sensor: an optical neuromorphic tactile sensor which aims to replicate spike-based human tactile transduction with an event-based approach.}

\begin{figure}[t]
	\centering
	\vspace{-1em}
	\includegraphics[width=0.9\linewidth]{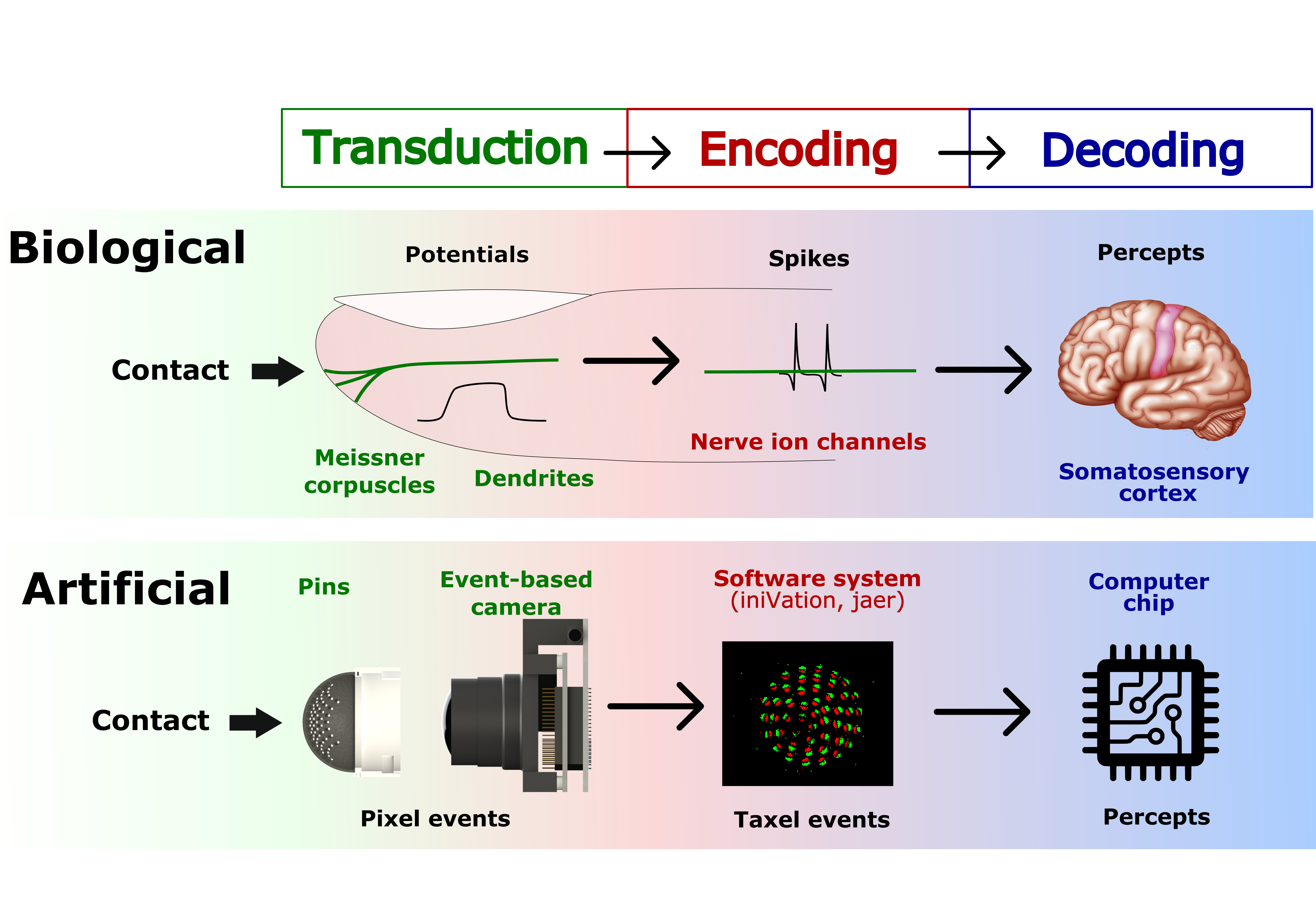}
	\vspace{-0.5em}
	\caption{Transduction, encoding and decoding mechanisms for the neuroTac sensor. The sensor mimics biological processes by accumulating pixel events (Potentials) from an event-based camera and combining them into taxel events (Spikes).}		
	\vspace{-1.5em}
	\label{fig:biomimetic_touch}
\end{figure}

The neuroTac sensor, described here, follows the tradition of neuromorphic technologies~\cite{liu2010neuromorphic} in seeking to produce and decode spike-based information (Fig.~\ref{fig:biomimetic_touch}). The spike-based sensor output is coded using 4 different methods: intensive (overall number of spikes), spatial (number of spikes per taxel), temporal (number of spikes per time window) and spatiotemporal (Van Rossum metric~\cite{rossum2001novel}). Although neuromorphic devices may present certain advantages over their non-neuromorphic counterparts (speed and energy efficiency), our principal objective here is linked to biomimetism. The TacTip sensor emulates the internal structure of human skin~\cite{ward2018tactip}, and the neuroTac adds to that biomimetic morphology by producing a neuromorphic, spike-based output. We aim to develop the neuroTac and associated spike-based information processing methods to investigate the advantages this approach might provide biological organisms.

The performance of the sensor and its associated coding methods is validated on texture recognition tasks, in which artificial and natural textures are identified using a K-nearest-neighbour (KNN) algorithm. The sensor successfully performs texture recognition, with temporal methods producing the highest classification accuracy. This  underlines the importance of spike timing in texture recognition tasks. 

%In this article we present the neuroTac, a novel, neuromorphic optical tactile sensor. We describe the biomimetic algorithm which combines event-based pixel-level information into taxel-level events analogous to spikes in the human nervous system. Finally, we validate the sensor's capabilities on a texture recognition task, in which we assess four different encoding mechanisms and compare their effectiveness. For this task, we found that a spatiotemporal encoding which accounts for both spike timing and sensor topology performs best when compared with intensive, spatial or temporal encodings. \notes{sth about speed?}.

%%%%%%%%%%%%%%%%%%%%%%%%%%%%%%%%%%%%%%%%%%%%%%%%%%%%%%%%%%%%%%%%%%%%%%%%%%%%%%%%%%%%%%%%%%%%%%%%%%%%%%%%
\section{BACKGROUND AND RELATED WORK}
\label{sec:background}

One of the strongest motivations for studying the human sense of touch is its important role in our interactions and in the manipulation of our environment. Artificial tactile sensing thus often mimics particular features of biological touch through biomimetic hardware or perception algorithms~\cite{dahiya2010tactile}.
An emerging area of biomimetic tactile sensing aims to replicate biological spike-based signalling through event-based systems and study the encoding of tactile information~\cite{johansson2009coding}. This is generally referred to as neuromorphic sensing~\cite{liu2010neuromorphic}.

Neuromorphic sensing arose from seminal work on silicon neurons~\cite{indiveri2011neuromorphic} and in the area of vision it has led to the successful integration of artificial retinas with the human nervous system~\cite{zaghloul2006silicon}. Evidence suggests that a neuromorphic tactile sensor could similarly be used to restore a natural sense of touch to amputees~\cite{raspopovic2014restoring,osborn2018prosthesis}.

Large-scale event-based tactile sensors have been developed for use as robotic skins~\cite{bergner2015event}, with a focus on the efficient processing of large quantities of asynchronous data. Bartolozzi et. al. also developed an architecture for use with distributed off-the shelf tactile sensors, transforming raw data into event-based signals~\cite{bartolozzi2017event}. Another example of a large-scale system for the investigation of spiking outputs in neuromorphic touch is that developed by Lee et. al.~\cite{lee2015kilohertz}. These systems represent crucial technological developments in the area of event-based systems, and will be essential in the creation of rapidly reacting fully tactile robot systems.

Here, our focus is not on large-scale event-based robotic skins or platforms for investigating spike-based encoding but rather on an artificial fingertip with high spatial resolution for fine in-hand manipulation tasks. Oddo et. al. developed a tactile sensor more in line with this design objective~\cite{oddo2009artificial}. Their neuromorphic tactile fingertip comprises 16 taxels combining raw outputs from 4 microelectromechanical system (MEMS) sensors, fed to an Izhikevich model to produce spikes. This system simulates the outputs from biological SA1 afferents, and has been proven capable of accurately distinguishing natural textures~\cite{rongala2015neuromorphic}. A crucial difference with the sensor presented here is that the neuroTac emulates the fast-adapting responses of FA1 afferents to dynamic contact (rather than the slowly adapting SA1 afferent responses). Recently, a sensor has been developed using similar hardware, with an event-based camera capturing the deformations of a deformable membrane~\cite{naeini2019novel}. The system was calibrated for use as a force sensor and to measure material hardness. The neuroTac sensor design aligns more closely both with traditional tactile sensor designs comprising taxels (internal markers which transduce contact), and biological fingertip structures (with taxels functioning as artificial mechanoreceptors). 

We validate the neuroTac on a texture classification task, in which 3d-printed and natural textures are classified by sliding the sensor horizontally across each stimulus. Texture identification is a common task in tactile sensing due to its important implications in object recognition and manipulation. Studies on the tactile sensing of textures generally involve naturally occurring  textures~\cite{jamali2011majority,rongala2015neuromorphic,weber2013spatial,xu2013tactile}. Here, we initially wish to investigate the sensor's response to highly structured textures. To achieve this, we utilize purposely designed 3d-printed stimuli with regular grids of cylindrical bumps as has been done in past studies of biological touch~\cite{srinivasan1990tactile}. Following this, we evaluate the sensor's classification performance on a set of 20 natural textures. 

%%%%%%%%%%%%%%%%%%%%%%%%%%%%%%%%%%%%%%%%%%%%%%%%%%%%%%%%%%%%%%%%%%%%%%%%%%%%%%%%%%%%%%%%%%%%%%%%%%%%%%%%
\section{METHODS}

\subsection{NeuroTac design and operation}
\label{sec:NeuroTacOperation}

\begin{figure}[t]
	\centering
	\includegraphics[width=0.75\linewidth]{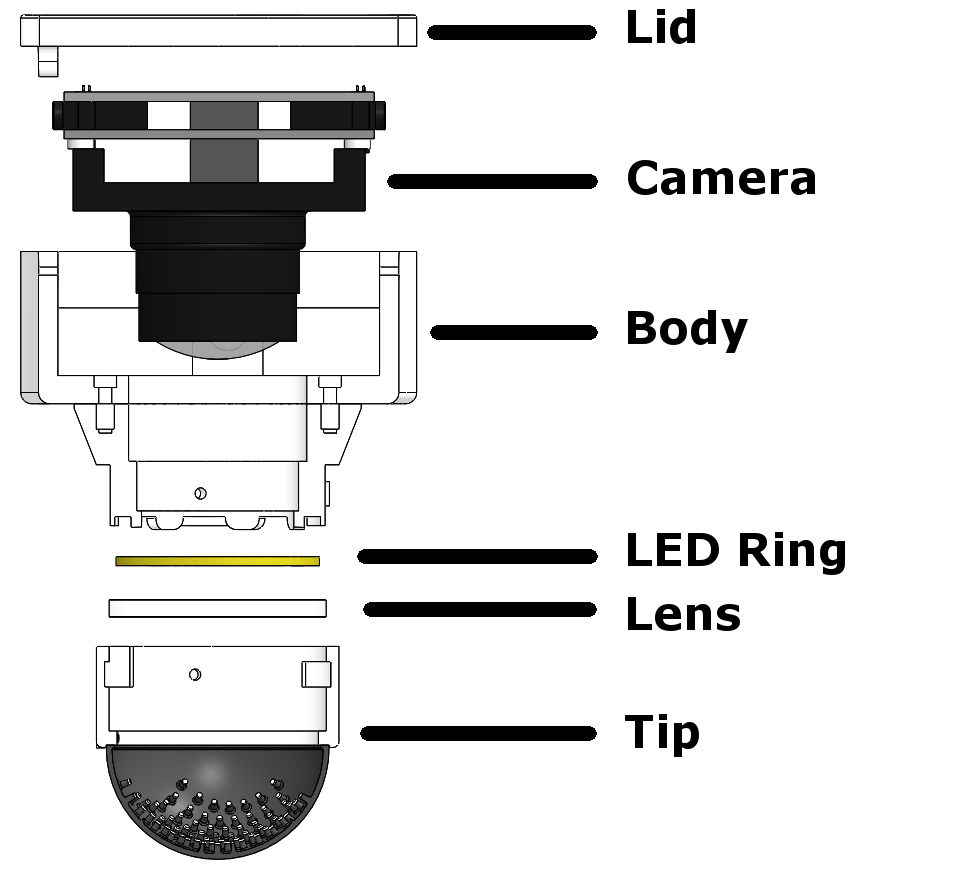}
	\caption{The neuroTac sensor. The tip contains internal pins treated as mechanoreceptors, which produce pixel events at the event-based camera. These are pooled and converted to taxel events (akin to biological spikes) upstream.}		
	\vspace{-1em}
	\label{fig:neuroTac_CAD}
\end{figure}

\paragraph{Sensor design}
The NeuroTac is based on the TacTip sensor~\cite{ward2018tactip}, a 3d-printed optical tactile sensor with a compliant dome-shaped outer membrane comprising biomimetic internal markers which emulate the internal structure of human fingertips~\cite{chorley2009development}. To convert the TacTip into a neuromorphic sensor, we replace its camera module (ELP, USBFHD01M-L21) with an event-based camera (iniVation, DAVIS240) which processes only dynamic events within the image frame. The DAVIS240 is an evolution of the DVS128~\cite{lichtsteiner2008128} and comprises 240x180 pixels, which process changes in brightness through independent electronic circuits to produce events in the address-event representation (AER) format~\cite{brandli2014240}. These events are combined by taxel and transmitted by the sensor, analogously to biological spike trains (see Section~\ref{sec:NeuroTacOperation} for more details on the sensor operation).

Like the TacTip, the NeuroTac is made up of 3 main hardware elements (Fig.~\ref{fig:neuroTac_CAD}): 

\begin{itemize}
	\item Tip: This is a compliant, 3d-printed modular part whose outer membrane comes into contact with the environment. Its internal surface comprises white-tipped pins which are displaced during contact. It is filled with silicone gel (Techsil, RTV27905) and covered with an acrylic lens to protect electronic components. 
	\item LED ring: This illuminates the tip's internal surface. 
	\item Camera: Housed within the main part of the sensor (the 3d-printed Body), this camera is event-based (iniVation, DAVIS240) and produces AER outputs in response to movements of the sensor's internal markers. 
\end{itemize}

\begin{figure}[t]
	\centering
	\includegraphics[width=0.9\linewidth]{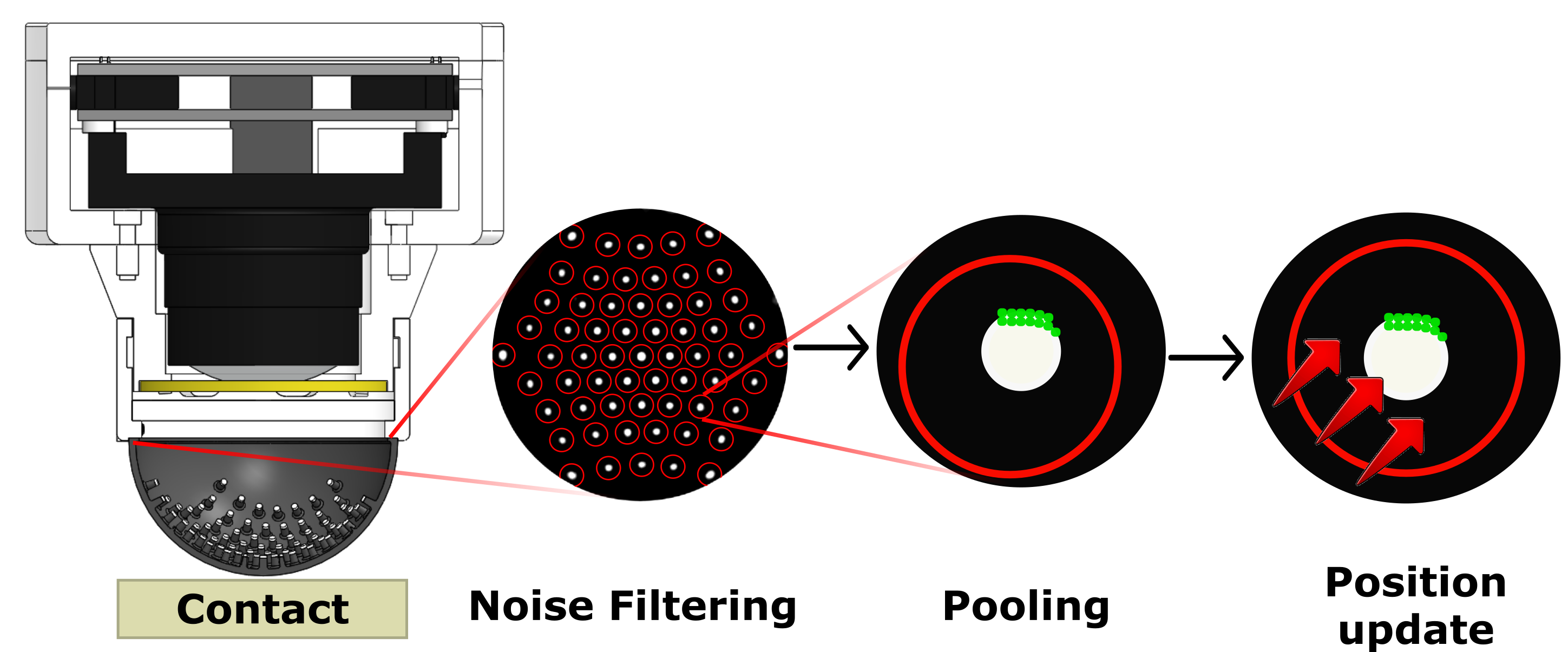}
	\vspace{1em}
	\caption{Sensor operation. Pixel events produced by the camera (iniVation, DAVIS240) are initially filtered, then pooled into a single taxel event. Finally, the position of taxels is updated (see Section~\ref{sec:methodsOperation}).}		
	\vspace{-0.5em}
	\label{fig:pixelevents}
\end{figure}

\begin{figure}[t]
	\centering
	\includegraphics[width=0.9\linewidth]{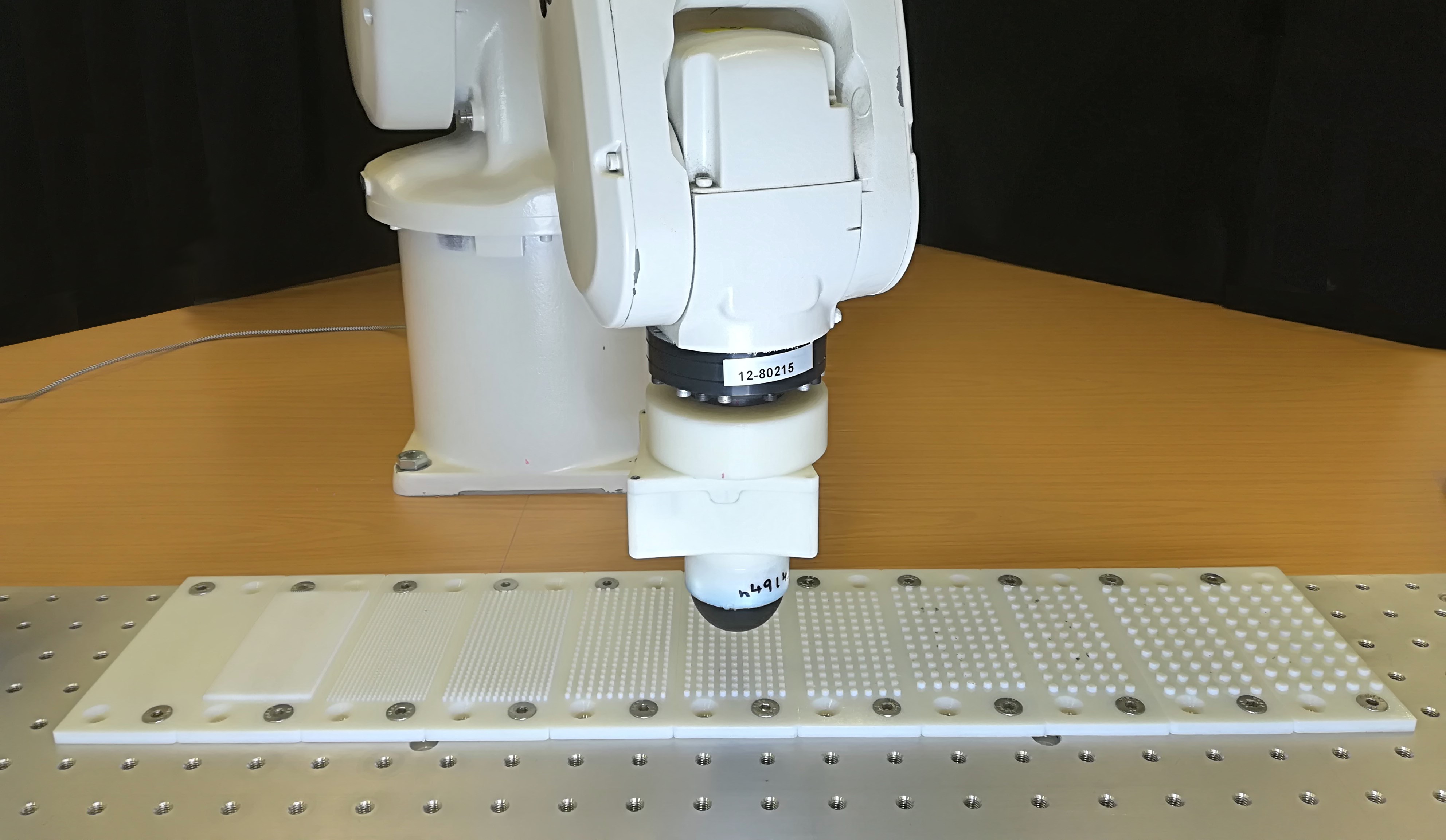}
	\caption{Experimental setup: The NeuroTac is attached to a 6-dof industrial robot arm (ABB, IRB120) which is used to slide the sensor horizontally across the 3d-printed textures.}		
	\vspace{-1.5em}
	\label{fig:ABBsetup}
\end{figure}

\paragraph{Sensor operation}
\label{sec:methodsOperation}

As the NeuroTac's compliant membrane deforms through contact, its 49 white-tipped internal pins deflect, and their movement triggers events in the camera (iniVation, Davis240) in the Address-Event Representation (AER) format. These events are produced by thresholding brightness changes at each photodiode in parallel~\cite{lichtsteiner2008128,brandli2014240}, leading to fast data transmission and high temporal precision. We designate these events 'pixel events' as they are created at the pixel level. Data transduction by the neuroTac then occurs through the 3 following steps (illustrated in Fig.~\ref{fig:pixelevents}):

\begin{itemize}
\item Noise filtering: each of the sensor's 49 pins represents a taxel, and has a receptive field assigned to it (6 px diameter). Pixel events that occur outside a taxel's receptive field, or which do not have another pixel event occur within a given spatiotemporal window (neighbouring pixels, 5\,ms) are filtered out. 
\item Pooling: pixel events are pooled over a short duration (20\,ms) and combined into a 'taxel event' based on the receptive field they are located within. Each taxel event is an array comprising 3 numbers: the number of pixel events it contains, their average location and their average timing. Note that each taxel event is interpreted as a spike associated with an artificial mechanoreceptor.
\item Position update: Receptive fields are re-centred around each taxel by shifting towards detected pixel events, to account for each pin's movement across the image.
\end{itemize}

% $x = x+ m\times \Delta x$, and $y = y+ m\times \Delta y$ ($\Delta x, \Delta y$  are the distances between the taxel and taxel event, and m is a gain factor set to )

%\begin{figure}[t]
%	\centering
%	\includegraphics[width=\linewidth]{bloc_diagram}
%	\caption{\notes{include this?}}		
%	\vspace{-0.5em}
%	\label{fig:bloc_diagram}
%\end{figure}

\subsection{Experimental setup}
\label{sec:exp_setup}

The first experiment with artificial textures involves sliding the NeuroTac across 11 3d-printed textures. Artificial textures consist of rectangular grids of cylindrical bumps (1\,mm height) with equal spacing and diameter. The texture grid size varies from 0\,mm (smooth surface) to 5\,mm in steps of 0.5\,mm (Fig.~\ref{fig:ABBsetup}).

%\subsubsection{Data collection}
%\label{sec:data_collection}

%\begin{figure}[t]
%	\centering
%	\includegraphics[width=\linewidth]{textures/AllStimuliv1}
%	\caption{3d-printed textures used during the classification experiment. Texture grid size varies from 0\,mm (Left) to 5\,mm (Right) in 0.5\,mm increments}		
%%	\vspace{-0.5em}
%	\label{fig:textures}
%\end{figure}

\begin{figure}[t]
	\centering
	\includegraphics[width=0.9\linewidth]{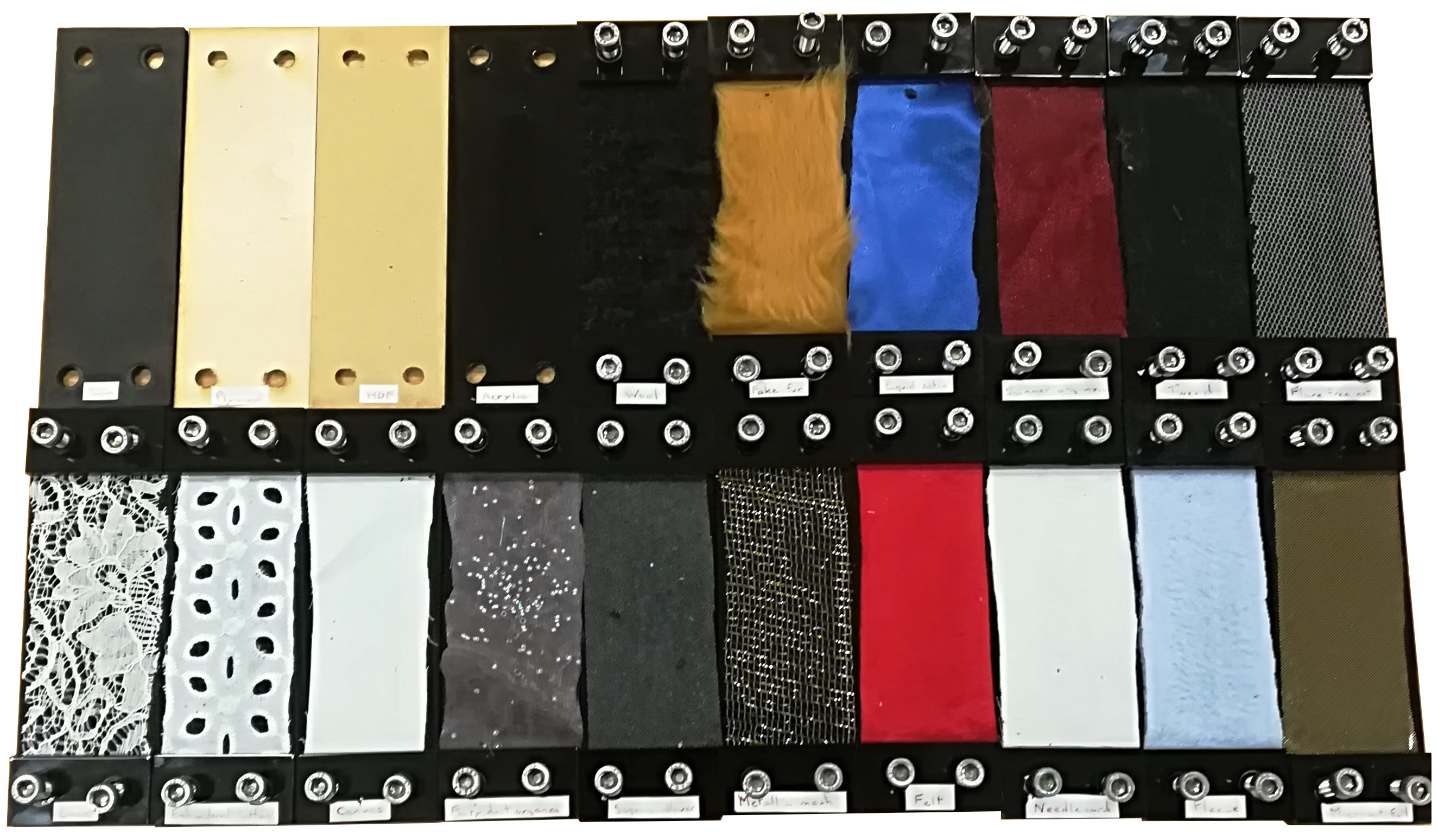}
	\caption{Natural textures used for classification. 20 textures were used, with a full list presented in Table~\ref{tab:natural_textures}.}		
	\vspace{-0.5em}
	\label{fig:natural_textures_setup}
\end{figure}

%\subsubsection{Data collection}
%\label{sec:data_collection}

Data is collected by mounting the NeuroTac on a 6-dof robotic arm (ABB, IRB120) and sliding it horizontally across the textures (Fig.~\ref{fig:ABBsetup}). 
The robot slides the NeuroTac across each texture at a speed of 15\,mm/s, over a distance of 60\,mm, comprising one data sample. We collect 100 samples for each texture, to obtain a dataset of 100 (number of runs) $\times$ 11 (number of textures) $\times$  49 (number of taxels) spike trains. 

In the second experiment on natural texture classification, the data collection procedure is repeated for 50 runs over 20 natural textures(see Table~\ref{tab:natural_textures}).
%The information contained within these spike trains can be encoded in different ways to represent the contact information characteristic of each texture. In the following, we describe 4 encoding methods which we apply to the texture classification problem in combination with a K-Nearest-Neighbours classifier to identify the textures. 

\subsection{Spike train encoding methods}
\label{sec:methods}

The multi-taxel spike trains obtained from the sensor are denoted as a matrix of spike times $t_n^i$, where $n = 1,2...,N$ (N denotes the number of taxels, $N=49$ in this case) and $i = 1... I_n$ ($I_n$ denotes the number of spikes for taxel $n$). A sample is denoted as the multi-taxel spike train resulting from the sensor sliding 60\,mm horizontally across one texture.
Following, we describe the 4 coding methods applied to transform the spike trains $t_n^i$ to the encoded representation $R$.

\subsubsection{Intensive coding}
\label{sec:methods_intensive}
This encoding consists of the average spike count per taxel for a given data sample. We name it intensive, analogously to work on biological touch~\cite{miyaoka1999mechanisms}, since it can be interpreted as an overall intensity of the sample signal with an absence of any spatial or temporal resolution. The encoding produces a single average spike count per sample, which is used for texture classification.

\begin{equation}
R = \frac{1}{N}\sum_{n=1}^{N} \sum_{i=1}^{I_n} t_n^i
\end{equation}
where $N$ is the total number of taxels and $I_n$ denotes the number of spikes for taxel n.

\subsubsection{Spatial coding}
\label{sec:methods_spatial}
Here we consider the topology of the sensor, and sum the spike counts separately for each of the sensor's 49 taxels. The resulting array of 49 spike rates is used as an encoded representation of the texture

\begin{equation}
R_n = \sum_{i=1}^{I_n} t_n^i
\end{equation}

\subsubsection{Temporal coding}
\label{sec:methods_temporal}
Temporal coding considers a rolling window of width $\Delta t$ over each data sample, which is rolled forward over the time domain in timesteps of 1\,ms. Within the window  $\Delta t$, the average spikes per taxel are recorded before proceeding to the next timestep. 

\begin{equation}
R_n(t) = \frac{1}{N} \sum_{t = t}^{t+\Delta t} t_n^i(t)
\end{equation}
\\
The $\Delta t$ parameter in this encoding method affects the classification accuracy, therefore we optimize it through brute-force optimization within a limited range of 1-200\,ms (see Section~\ref{sec:inspection}).

\subsubsection{Spatiotemporal coding}
\label{sec:methods_spatiotemporal}

Spatiotemporal coding uses the spatial and temporal features of spike trains produced by the sensor. A multi-neuron Van Rossum distance~\cite{houghton2008new} is used as a metric in the texture classification, to ensure that both spatial and temporal dimensions of the data are considered. 

The spatiotemporal encoded representation can be considered the convolution of the multi-taxel spike train with an exponential kernel used in the Van Rossum distance calculation (see Section~\ref{sec:methods_classification}).

\begin{equation}
R_n^i(t) = t_n^i h(t-t_i)
\end{equation}

where
\begin{equation}
h = 
\begin{cases}
0, & t\le 0\\
\frac{1}{\tau}e^{-t/\tau},              & t \geq 0
\end{cases}
\end{equation}
$\tau$ is a time constant parameter which can be optimized to create the most accurate clustering of texture representations.

\subsection{Classification}
\label{sec:methods_classification}

Texture classification is performed through a KNN algorithm (k=4), which assigns a class (texture grid size 0-5\,mm in 0.5\,mm steps here) to a test sample based on its 4 closest training samples. 
For the intensive, spatial and temporal coding methods, a standard Euclidean distance is used to calculate distances between samples. 

For the spatiotemporal coding, we use a multi-neuron Van Rossum distance~\cite{houghton2008new} which involves the convolution of spike trains with an exponential kernel (Section~\ref{sec:methods_spatiotemporal}) before applying a distance metric.

In the original Van Rossum metric~\cite{rossum2001novel}, which applies to single neuron spike trains, the distance metric between two spike trains is simply:

%The Van Rossum metric~\cite{rossum2001novel} compares two spike trains by converting both spike trains to continuous functions through the use of a kernel $h(t)$:
%\begin{equation}
%s \rightarrow f(t;\mathbf{t}) = \sum_{i=1}^n h(t-t_i)
%\end{equation}
%The distance between the two spike trains can then be calculated with the following formula: 

\begin{equation}
d_2(t_1,t_2;h)= \sqrt{\int dt (f_1 - f_2)^2}
\end{equation}
where $f_1$ and $f_2$ are the convolved functions.

The extension of the Van Rossum metric to multi-neuron cases, as described by Houghton and Sen~\cite{houghton2008new}, introduces an additional parameter $\theta$ which represents the correlation between neurons. The parameter's effect is best described by taking its cosine, which varies from 0 to 1 with:

\begin{itemize}
	\item  $cos\theta = 0$. Labelled line code: each neuron is considered independent and their distances are summed.
	\item $cos\theta = 1$. Summed population code: spike trains are superimposed before calculating the distance metric. 
	\item $1 > cos\theta > 0$: an intermediate level of inter-neuron correlation between these two extremes.
\end{itemize}
 
The $\theta$ angle is an optimizable parameter, and thus we perform a two parameter Bayesian optimization of $\tau$ (exponential kernel time constant) and $\theta$ for spatiotemporal encoding as described in section~\ref{sec:artificialClassification}.

% Once again, we refer the reader to~\cite{houghton2008new} for a more detailed, mathematical account of the multi-neuron Van Rossum metric.

%%%%%%%%%%%%%%%%%%%%%%%%%%%%%%%%%%%%%%%%%%%%%%%%%%%%%%%%%%%%%%%%%%%%%%%%%%%%%%%%%%%%%%%%%%%%%%%%%%%%%%%%
\section{RESULTS}

%Times : 
%5mm/s : 13.177s
%10 mm/s: 6.688s
%15 mm/s: 4.528s
%20 mm/s: 3.439s
%25 mm/s: 2.784s

\subsection{Inspection of data - Artificial textures}
\label{sec:inspection}

\begin{figure}[t]
	\centering
	\begin{subfigure}[b]{0.32\linewidth}
		\centering
		\includegraphics[width=\linewidth]{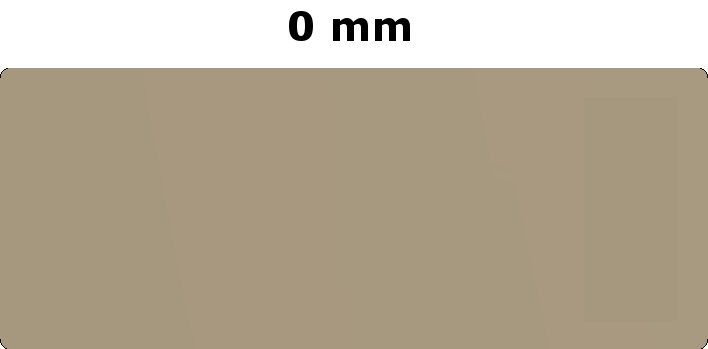}
	\end{subfigure}
	\begin{subfigure}[b]{0.32\linewidth}
		\centering
		\includegraphics[width=\linewidth]{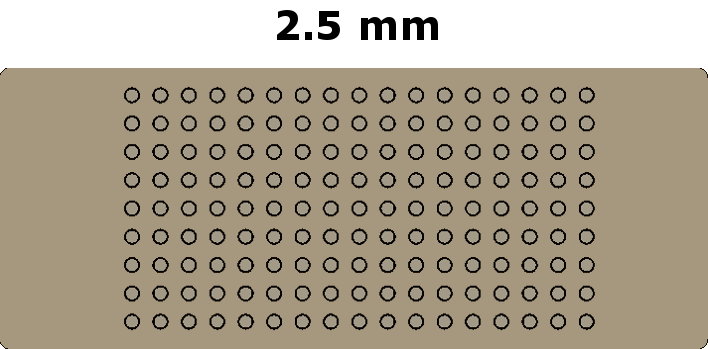}
	\end{subfigure}
	\begin{subfigure}[b]{0.32\linewidth}
		\centering
		\includegraphics[width=\linewidth]{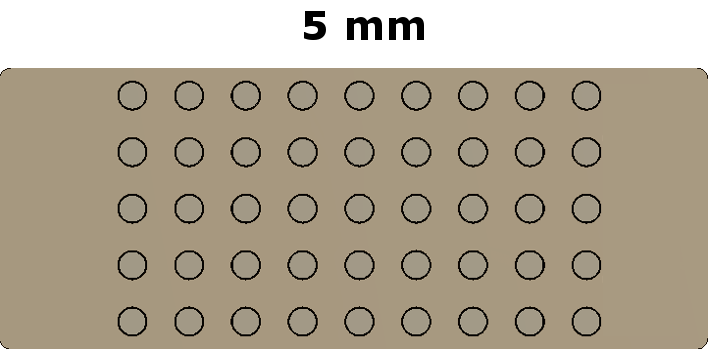}
	\end{subfigure}
	\begin{subfigure}[b]{0.32\linewidth}
		\centering
		\includegraphics[width=\linewidth]{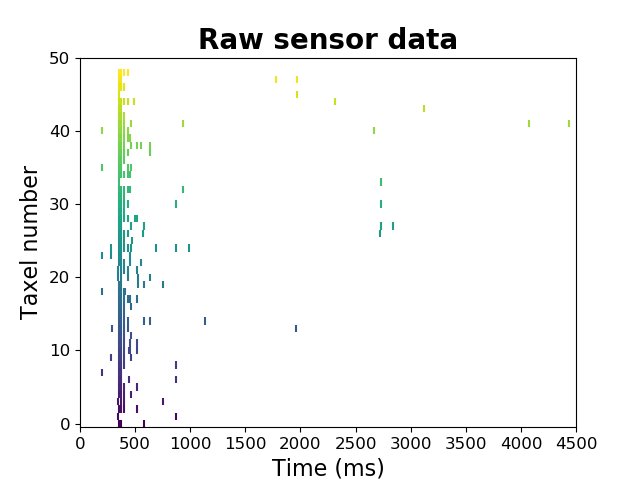}
	\end{subfigure}
	\begin{subfigure}[b]{0.32\linewidth}
		\centering
		\includegraphics[width=\linewidth]{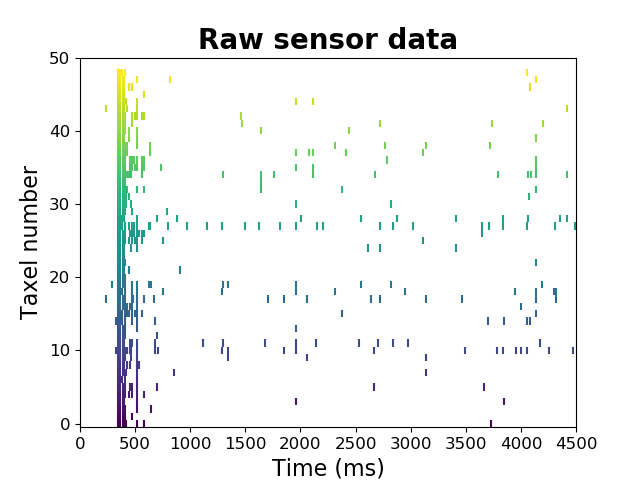}
	\end{subfigure}
	\begin{subfigure}[b]{0.32\linewidth}
		\centering
		\includegraphics[width=\linewidth]{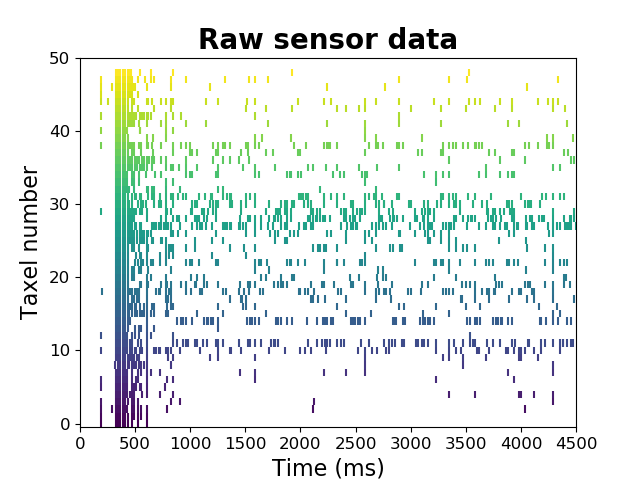}
	\end{subfigure}
	\begin{subfigure}[b]{0.32\linewidth}
		\centering
		\includegraphics[width=\linewidth]{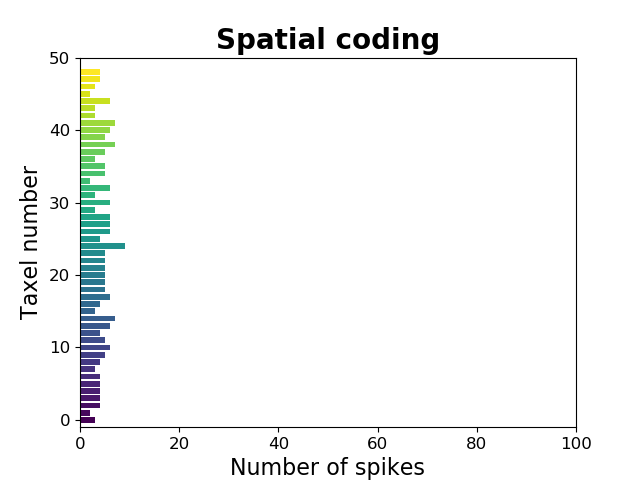}
	\end{subfigure}
	\begin{subfigure}[b]{0.32\linewidth}
		\centering
		\includegraphics[width=\linewidth]{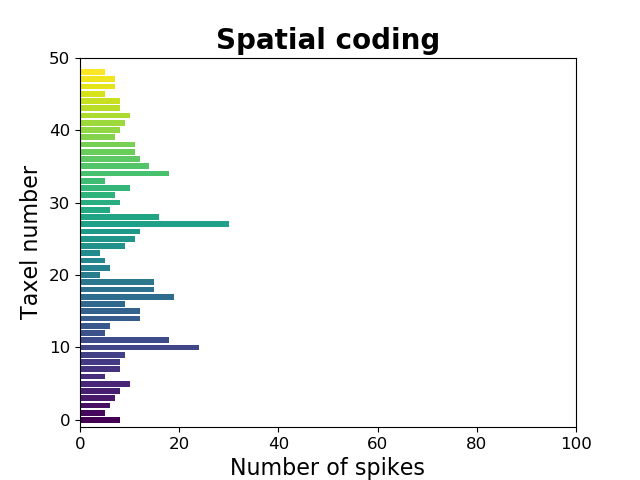}
	\end{subfigure}
	\begin{subfigure}[b]{0.32\linewidth}
		\centering
		\includegraphics[width=\linewidth]{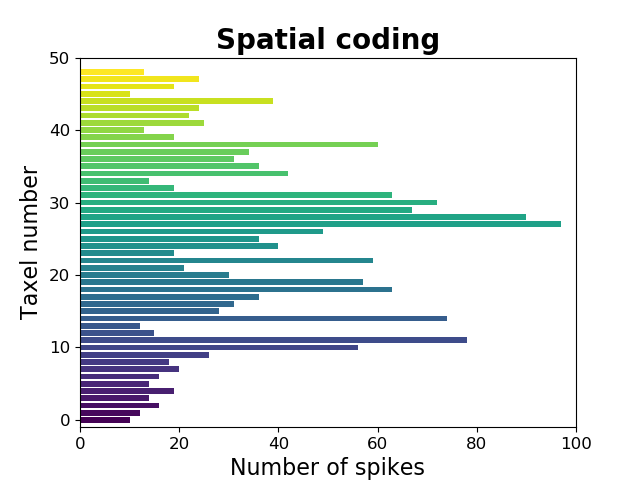}
	\end{subfigure}
	\begin{subfigure}[b]{0.32\linewidth}
		\centering
		\includegraphics[width=\linewidth]{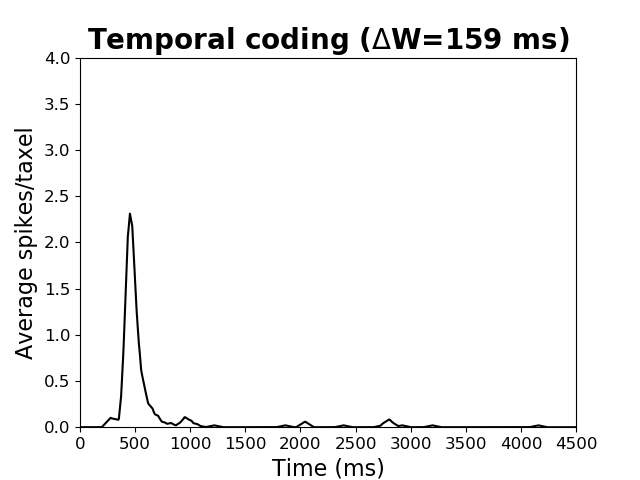}
	\end{subfigure}
	\begin{subfigure}[b]{0.32\linewidth}
		\centering
		\includegraphics[width=\linewidth]{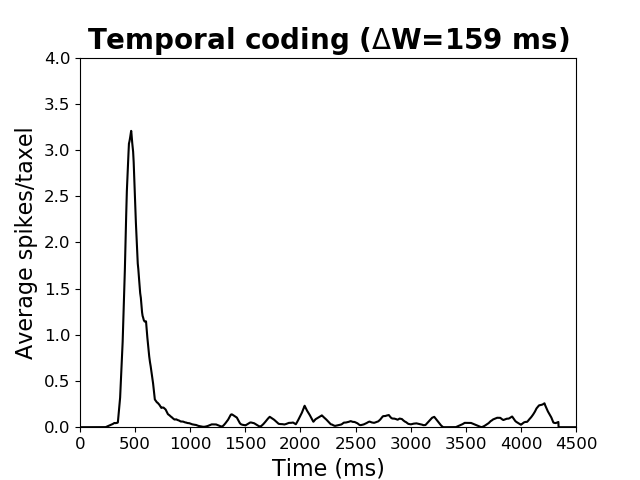}
	\end{subfigure}
	\begin{subfigure}[b]{0.32\linewidth}
		\centering
		\includegraphics[width=\linewidth]{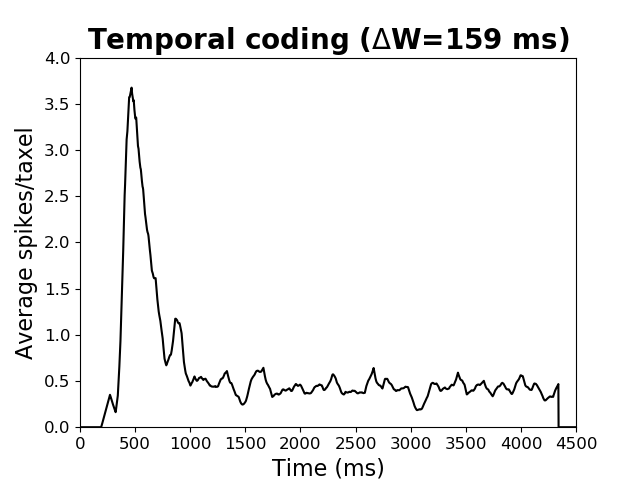}
	\end{subfigure}
	\caption{Examples of the spike trains produced by the NeuroTac when sliding across three different 3d-printed textures. From left to right, textures are 0\,mm grid (smooth), 2.5\,mm grid and 5\,mm grid, with the corresponding spike trains, spatial and temporal distributions displayed below them. }
	\label{fig:spikes_textures}
	\vspace{-0.25em}
\end{figure}

Data is gathered by sliding the sensor horizontally across the set of 11 artificial textures (see Section~\ref{sec:exp_setup}). The output of the NeuroTac consists of 49 spike trains which represent the events being produced at each taxel. The spike trains will vary in overall intensity for different textures, as well as having distinct spatial and temporal signatures.

Examples of spike trains obtained for 3 textures of grid sizes 0\,mm (smooth), 2.5\,mm and 5\,mm are displayed here (Fig.~\ref{fig:spikes_textures}, second row). It is visually noticeable that these 3 multi-neuron spike trainsdiffer, with the number of spikes produced increasing with texture coarseness. As expected, applying intensive coding to these samples gives readily distinguishable average spike counts over all taxels of $4.63$ spikes/taxel (0\,mm grid), $9.84$ spikes/taxel  (2.5\,mm grid) and $34.73$ spikes/taxel (5\,mm grid).

Spatial coding (spike frequency per taxel) reveals an additional topological structure to the data (Fig.~\ref{fig:spikes_textures}, third row), wherein certain taxels seem to fire at higher rates than others (taxels 10, 17 and 27 for the 2.5\,mm texture for instance). This could be due to their location on the sensor being directly in the path of the texture's raised bumps, leading to more events being produced for those taxels. 

We also illustrate temporal coding with a rolling window of size $\Delta t = 159\,ms$ (Fig.~\ref{fig:spikes_textures}, bottom row), in which the increased response with texture grid size is visibly apparent. There is also a sharp increase in activity at the beginning of each sample which corresponds to the start of the sensor's horizontal sliding motion, and is likely linked to the coefficient of static friction between the sensor and texture.
% This could provide an additional feature for texture classification.

\subsection{Texture classification - Artificial textures}
\label{sec:artificialClassification}

\begin{figure}[t]
	\centering
	\begin{subfigure}[t]{0.49\linewidth}
		\includegraphics[width=\linewidth]{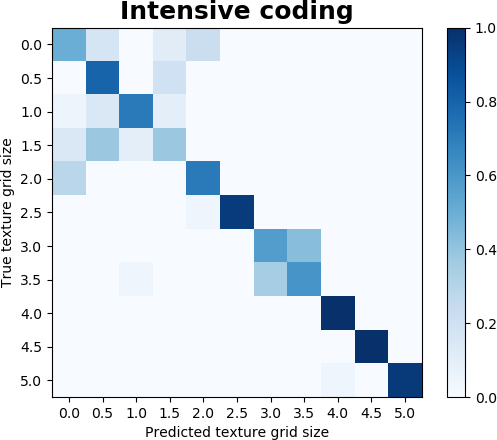}
	\end{subfigure}
	\begin{subfigure}[t]{0.49\linewidth}
		\includegraphics[width=\linewidth]{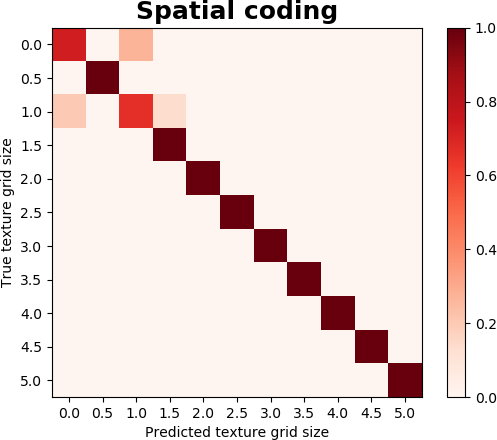}
	\end{subfigure}
%\vspace{1em}
	\begin{subfigure}[t]{0.49\linewidth}
		\includegraphics[width=\linewidth]{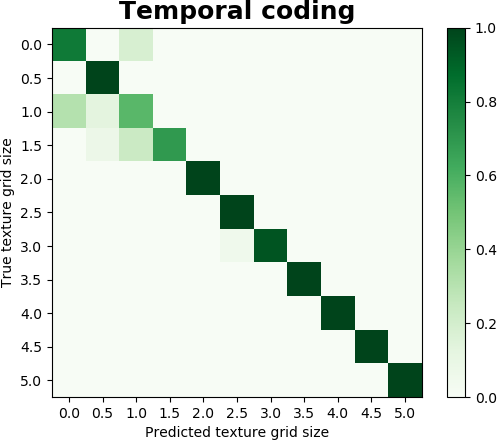}
	\end{subfigure}
	\begin{subfigure}[t]{0.49\linewidth}
		\includegraphics[width=\linewidth]{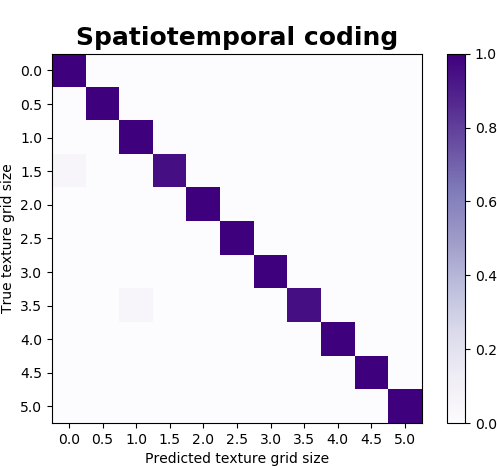}
	\end{subfigure}
	\vspace{-1em}
	\caption{Confusion matrices for artificial texture classification with each of the 4 encoding methods: Intensive (Top Left), Spatial (Top Right), Temporal (Bottom Left) and Spatiotemporal (Bottom Right).}	
	\label{fig:confusion_matrices}
	\vspace{-1em}
\end{figure}

For each encoding method (see Section~\ref{sec:methods}), the resulting data is classified with a KNN algorithm to identify the 11 3d-printed textures. 

Note that temporal and spatiotemporal encodings are both parameterised (see Section~\ref{sec:methods_temporal},~\ref{sec:methods_spatiotemporal}), and their parameters ($\Delta t$ for temporal encoding, $cos\theta$ and $\tau$ for spatiotemporal encoding) are optimised to maximize classification accuracy. Temporal encoding contains a single parameter $\Delta t$, allowing us to perform a brute-force optimization over a range of values from 1 to 200\,ms. We set an upper threshold on the $\Delta t$ parameter both to accelerate the optimization procedure and ensure the method is distinct from intensive coding (which corresponds to $\Delta t = T$, where T is the full duration of the sample). This optimization process give us a value of $\Delta t = 159\,ms$.

In the case of the spatiotemporal encoding, we perform a Bayesian optimization over the two parameters  $cos\theta$ and $\tau$. The domain space we apply optimization over is $cos \theta = 0-1$ and $\tau = 10-100 ms$. We place an upper limit on the $\tau$ parameter for faster convergence and to ensure spatiotemporal coding is distinct from spatial coding. The parameters converge over 1000 optimization epochs to values of $cos\theta = 0.4$ and $\tau = 76 ms$.

%\begin{figure}[t]
%	\centering
%\begin{subfigure}[b]{0.32\linewidth}
%	\centering
%	\includegraphics[width=\linewidth]{textures/TemporalOptimizerBruteforceResults}
%\end{subfigure}
%\begin{subfigure}[b]{0.32\linewidth}
%	\centering
%	\includegraphics[width=\linewidth]{textures/SpatiotemporalOptimizerTau}
%\end{subfigure}
%\begin{subfigure}[b]{0.32\linewidth}
%	\centering
%	\includegraphics[width=\linewidth]{textures/SpatiotemporalOptimizerCosTheta}
%\end{subfigure}
%	\caption{Parameter optimization. Left: Brute-force optimization of $\Delta W$. A KNN classification algorithm runs with $\Delta W = 1-200\,ms$ in steps of 1\,ms, and $\Delta W = 159\,ms$ is chosen as the value with the highest classification accuracy. Middle: Convergence of $\tau$ over 1000 epochs of a Bayesian optimization procedure (final value: $\tau = 76\,ms$). Right: Convergence of $cos\theta$ over 1000 epochs of a Bayesian optimization procedure (final value: $cos\theta = 0.4$)}
%	\label{fig:parameters}
%	\vspace{-0.25em}
%\end{figure}

%\begin{figure}[t]
%	\centering
%	\includegraphics[width=0.7\linewidth]{textures/IntensiveValues}
%	\caption{Average spike counts per taxel (for all taxels and over the entire data sample) for all artificial textures. The values displayed here are averaged over 100 runs. These spike counts are used in the intensive coding method for texture classification.}		
%	\vspace{-0.5em}
%	\label{fig:intensive_coding}
%\end{figure}

\begin{table}[t]
  \begin{center}
    \begin{tabular}{l|c} % <-- Alignments: 1st column left, 2nd middle and 3rd right, with vertical lines in between
      \textbf{Coding method} & \textbf{Performance - Artificial textures (\%)} \\
      \hline
      \rule{0pt}{2ex}Intensive & 78.5 $\pm$ 41.1 \\
      Spatial & 95.5 $\pm$ 20.6\\
      Temporal & 98.1 $\pm$ 13.7\\
	  Spatiotemporal & 98.3 $\pm$ 13\\
    \end{tabular}
    \caption{Comparison of different encoding methods using leave-one-out cross-validation for texture classification.}
    \label{tab:table1}
  \end{center}
\vspace{-1.5em}
\end{table}

First, we run a simple 80/20 train-test split on the gathered dataset, and attempt to classify the textures using a KNN classifier (k=4). The results are presented in the form of confusion matrices (Fig.~\ref{fig:confusion_matrices}), and provide insight into each coding method's performance. Visually, we can observe that the intensive method is the least effective, particularly for the smoother textures (bump diameter 0-2\,mm). The spatial coding method seems more accurate, though there is still some slight inaccuracy in the classification of smoother textures (bump diameter 0-1.5\,mm). Temporal and spatiotemporal coding appears to provide near-perfect classification accuracy, indicating that sliding the sensor over the textures produces discriminable time-dependent features.

To obtain a more complete picture of the performance of each coding method, we perform a leave-one-out cross-validation over the gathered dataset. Results are presented in table~\ref{tab:table1} and mirror the results displayed in the train-test split confusion matrices. Intensive coding performs worst, with spatial information improving performance significantly. However both temporal and spatiotemporal coding have the highest accuracy, indicating the sensor produces features in the temporal domain which are likely most representative of the classified textures.   

%
%\begin{table}[h!]
%	\begin{center}
%		\begin{tabular}{l|c|c|r} % <-- Alignments: 1st column left, 2nd middle and 3rd right, with vertical lines in between
%			\textbf{Encoding} & \textbf{10-run class.} & \textbf{10-fold cross-val} & \textbf{Leave1out} \\
%			\hline
%			Intensive & 73.8 $\pm$ 2.4 & 76.4 $\pm$ 3.5 & 78.5 $\pm$ 41.1 \\
%			Spatial & 95.8 $\pm$ 1.3 & 95.9 $\pm$ 1.9 & 95.5 $\pm$ 20.6\\
%			Temporal & 98.5 $\pm$ 0.7 & 97.9 $\pm$ 1.3 & 98.1 $\pm$ 13.7\\
%			Spatiotemporal & $\pm$ & 95.1 $\pm$ 1.2 & 98.3 $\pm$ 13\\
%		\end{tabular}
%		\caption{Comparison of different encoding methods using k-fold cross-validation for texture classification.}
%		\label{tab:table1}
%	\end{center}
%\end{table}

\subsection{Texture classification - Natural textures}
\label{sec:results_natural}

Here we seek to test the sensor's performance more thoroughly on texture recognition by replacing the 3d-printed textures with natural textures, listed below in Table~\ref{tab:natural_textures}.

\begin{figure}[t]
	\centering
	\begin{subfigure}[t]{0.49\linewidth}
		\includegraphics[width=0.95\linewidth]{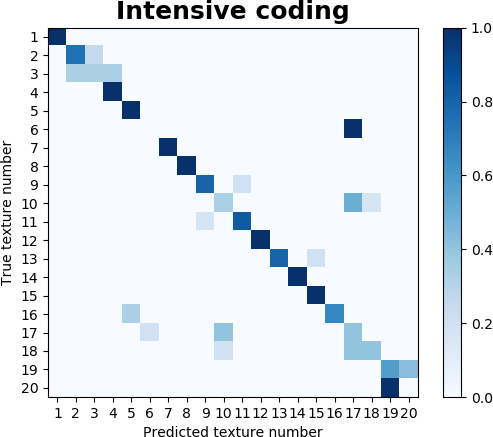}
	\end{subfigure}
	\begin{subfigure}[t]{0.49\linewidth}
		\includegraphics[width=0.95\linewidth]{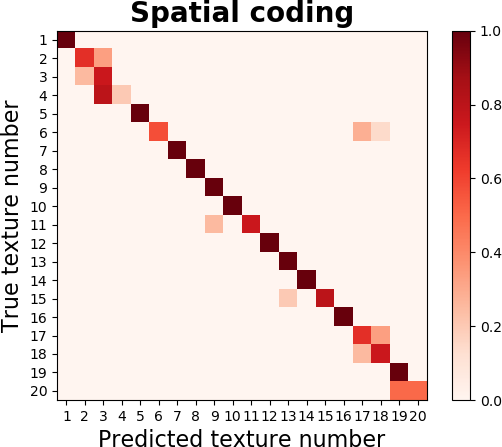}
	\end{subfigure}
	\par\medskip
	\begin{subfigure}[t]{0.49\linewidth}
		\includegraphics[width=0.95\linewidth]{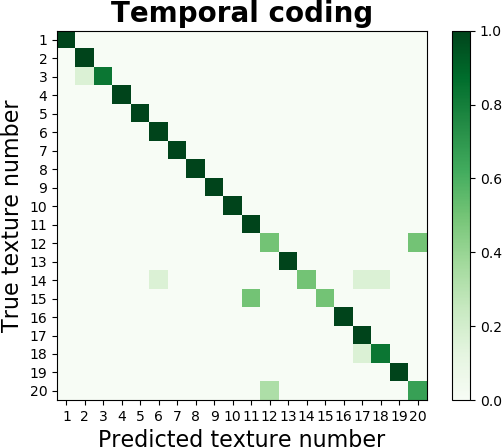}
	\end{subfigure}
	\begin{subfigure}[t]{0.49\linewidth}
		\includegraphics[width=0.95\linewidth]{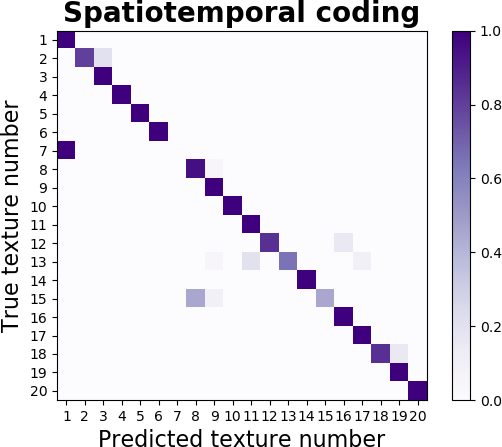}
	\end{subfigure}
	\vspace{-1em}
	\caption{Confusion matrices for natural texture classification with each of the 4 encoding methods: Intensive (Top Left), Spatial (Top Right), Temporal (Bottom Left) and Spatiotemporal (Bottom Right).}	
	\label{fig:confusion_matrices_natural}
	\vspace{-1em}
\end{figure}

We again run a KNN classifier (k=4) over 20 runs with an 80/20 train-test split  (giving 4 test runs for each texture). The results are presented in the form of confusion matrices for each coding method (Fig.~\ref{fig:confusion_matrices_natural}).

Classification accuracy appears generally strong, with Intensive coding the weakest method, and Temporal and Spatiotemporal coding performing the best, as was the case for artificial textures. 
This is confirmed through a leave-one-out cross-validation, with results displayed in Table~\ref{tab:results_natural}. 
One significant difference between the natural and artificial textures is that natural textures appear more irregular in their misclassifications. The artificial textures lie along a regularly designed gradient of coarseness, and thus classification errors occur most often between neighbouring classes. In the case of natural textures, there is no such regularity or order and classification errors appear to be spread more widely across the confusion matrices (Fig.~\ref{fig:confusion_matrices_natural}).

It is also interesting to note that classification errors vary between coding methods. For instance for Intensive and Spatial coding, fake fur and felt (textures 6 and 17) are confused, whereas Temporal and Spatiotemporal coding can distinguish them. Equally, liquid satin (texture 7) is misclassified as foam (texture 1) only when Spatiotemporal coding is applied (Fig.~\ref{fig:confusion_matrices_natural}). The fact that these misclassifications vary between coding methods likely stems from the fact that each method produces a distinct set of features.

\section{Discussion}

The neuroTac sensor developed here is a neuromorphic optical tactile sensor, with a data output consisting of multi-taxel spike trains. The spike trains were encoded using four different bio-inspired encoding mechanisms: Intensive, Spatial, Temporal and Spatiotemporal. We validated the sensor through a texture classification task, in which 11 3d-printed textures (grid sizes 0-5\,mm in steps of 0.5\,mm) and 20 natural textures were discriminated using a KNN classification algorithm. 

We chose texture discrimination as a validation task as it has been a key experimental procedure in psychophysical and neuroscientific studies of touch. Here we will attempt to contrast our results with existing theories in these fields, and identify limitations as well as gaps for further investigation. 

We found that applying spatial coding to the neuroTac data produced high classification accuracy for coarser textures (2.5-5\,mm grid size), but performed less well for smoother textures (0-2.5\,mm grid size). This result seems to concur with Katz's duplex theory of human tactile perception, with spatial resolution being important for rougher textures, and vibration cues taking over as textures get smoother~\cite{hollins2000evidence}. The stronger performance of the temporal and spatiotemporal coding methods for smooth textures could be linked to their detection of high frequency vibrational cues.

\begin{table}[t]
	\begin{center}
		\begin{tabular}{c|c|c|c}
			\textbf{Texture} & \textbf{Texture} & \textbf{Texture} & \textbf{Texture} \\ 
			\textbf{number} & \textbf{name} & \textbf{number} & \textbf{name} \\
			\hline		
			\rule{0pt}{2ex}1 & Foam & 11 & Flarefree net\\
			2 & Plywood & 12 & Embroidered cotton\\
			3 & MDF & 13 & Shirt canvas\\
			4 & Acrylic & 14 & Fairydust organza\\
			5 & Wool & 15 & Sequins allover \\
			6 &  Fake fur & 16 & Metallic mesh\\
			7 & Liquid satin & 17 & Felt\\
			8 & Shimmer organza & 18 & Needlecord\\
			9 & Tweed & 19 & Fleece\\
			10 & Lace & 20 & Microdot foil\\
		\end{tabular}
		\caption{Natural textures and their corresponding number.}
		\label{tab:natural_textures}
	\end{center}
\vspace{-1em}
\end{table}

\begin{table}[t]
	\begin{center}
		\begin{tabular}{c|c} % <-- Alignments: 1st column left, 2nd middle and 3rd right, with vertical lines in between
			\textbf{Coding method} & \textbf{Performance - natural textures (\%)} \\
			\hline
			\rule{0pt}{2ex}Intensive & 69.8 \\ 
			Spatial & 85.5\\
			Temporal & 93.0\\
			Spatiotemporal & 92.8\\
		\end{tabular}
		\caption{Comparison of the 4 different coding methods using leave-one-out cross-validation for texture classification of the 20 natural textures.}
		\label{tab:results_natural}
		\vspace{-1em}
	\end{center}
\vspace{-1em}
\end{table}

It has also been suggested in a recent study that the specific timing of spikes carries significance for texture discrimination, as it could encode spatial frequency features of the textures being contacted~\cite{mackevicius2012millisecond}. The spatiotemporal coding method described here could capture these precise spike timings, with a resolution dependent on the time constant parameter $\tau$. Here $\tau$ was optimized to be 76\,ms, which could indicate an approximate timescale for the frequency features of the textures considered.

%, demonstrating for instance the importance of spatial encoding relative to intensive coding on tasks such as grating orientation discrimination~\cite{phillips1981tactile}. 

%It is important to note however that texture classification performance seems to depend mostly on the signal obtained during the inception of motion of the sensor. During sliding, spike signals become very sparse for most textures and classification performance degrades, which is likely due to vibrations being damped by the sensor's 3d-printed skin and internal silicone gel. Texture identification during sliding could be improved by replacing these materials with alternatives with lower damping coefficients (such as a silicone moulded skin).

Human touch and proprioception provide much richer and more complex data than that explored here, and most theories suggest that human texture recognition relies on a complex coding strategy involving inputs from several mechanoreceptors (Pacinian corpuscles, Merkel cells)~\cite{mackevicius2012millisecond}. Further studies with the neuroTac could include methods for simulating input from these biological afferents, for instance by feeding taxel positions as an input to spiking neuron models. This would open the path to more complex coding algorithms to be applied to the sensor's neuromorphic data which could improve generalizability and robustness.

%%%%%%%%%%%%%%%%%%%%%%%%%%%%%%%%%%%%%%%%%%%%%%%%%%%%%%%%%%%%%%%%%%%%%%%%%%%%%%%%%%%%%%%%%%%%%%%%%%%%%%%%
\section{Conclusion}
We presented a neuromorphic optical tactile sensor and demonstrated its performance on a texture classification task. Four bio-inspired spike encoding mechanisms were investigated, which suggested information about texture coarseness is encoded in the timing of spikes. The neuroTac's fast spike-based output could lead to a step forward in the areas of robotic manipulation and prosthetics.

%{\em Acknowledgements:}

%%%%%%%%%%%%%%%%%%%%%%%%%%%%%%%%%%%%%%%%%%%%%%%%%%%%%%%%%%%%%%%%%%%%%%%%%%%%%%%%

\balance

\bibliographystyle{unsrt}

\bibliography{2019_RAL-ICRA_NeuroTac}

% revisit background

\end{document}